%% file: rios_tobar_ijcnn2018.tex
\documentclass[conference]{IEEEtran}

\pdfoutput=1
\usepackage{times} 
\usepackage{graphicx} 
\usepackage{subfigure}
\usepackage[utf8]{inputenc}

\usepackage{amsmath, amsthm, amssymb}

\include{formulas}
\graphicspath{{images/}}

\usepackage{algorithm}
\usepackage{algorithmic}
\usepackage{tabularx}

\title{Learning non-Gaussian Time Series using the Box-Cox Gaussian Process}
\author{
	\IEEEauthorblockN{Gonzalo Rios}
	\IEEEauthorblockA{Center for Mathematical Modeling and\\
					  Department of Mathematical Engineering\\
					  Universidad de Chile\\
					  grios@dim.uchile.cl}
\and
	\IEEEauthorblockN{Felipe Tobar}\thanks{mahi}
	\IEEEauthorblockA{Center for Mathematical Modeling\\
					  Universidad de Chile\\
					  ftobar@dim.uchile.cl}
}

\input{preamble}

\begin{document}
%
\maketitle
%
\input{abstract}
%

\input{intro}

\input{background}

\input{trans}
\input{exper}
\input{disc}
\section*{Acknowledgments}
This work was funded by Conicyt projects AFB 170001 (Center for Mathematical Modeling), Fondecyt \#11171165 (F.T.) and PCHA Doctorado Nacional 2016-21161789 (G.R.) 


\bibliographystyle{IEEEbib}
\bibliography{library}

\end{document}

%% file: preamble.tex
\usepackage{amsbsy}
\usepackage{amssymb}
\usepackage{amsmath,bm}
\usepackage{xcolor}

\def\GP{{\mathcal GP}}

\def\R{{\mathbb R}}

\def\cN{\mathcal{N}}

\def\y{\mathbf{y}}
\def\t{\mathbf{t}}

\def\x{\mathbf{x}}

\def\O{{\mathcal{O}}}

%% file: abstract.tex

\begin{abstract}
Gaussian processes (GPs) are Bayesian nonparametric generative models that provide interpretability of hyperparameters, admit closed-form expressions for training and inference, and are able to accurately represent uncertainty. To model general non-Gaussian data with complex correlation structure, GPs can be paired with an expressive covariance  kernel and then fed into a nonlinear transformation (or warping). However, overparametrising the kernel and the warping is known to, respectively, hinder gradient-based training and make the predictions computationally expensive. We remedy this issue by (i) training the model using derivative-free global-optimisation techniques so as to find meaningful maxima of the model likelihood, and (ii) proposing a warping function based on the celebrated Box-Cox transformation that requires minimal numerical approximations---unlike existing warped GP models. We validate the proposed approach by first  showing that predictions can be computed analytically, and then on a learning, reconstruction and forecasting experiment using real-world datasets.

\end{abstract}

%% file: intro.tex

\section{Introduction}
A Gaussian process (GP) \cite{rasmussen06} is a prior distribution over functions with a support that includes a wide class of phenomena via the design of its mean and covariance functions, the parameters of which provide meaningful interpretation of the process at hand. Beyond regression \cite{NIPS1995_1048}, GPs have been extensively used in the last two decades for classification \cite{NIPS1999_1694}, density estimation \cite{NIPS2008_3410}, filter design \cite{NIPS2015_5772}, model identification \cite{deisenroth_pami} and optimisation \cite{BayesianOpt}. In general terms, all these generative models have two stages: The latent process is modelled as a GP and the observation is modelled (conditional to the latent process) as a non-Gaussian variable. This class of models is referred to as GP with non-Gaussian likelihood, or as Generalised GPs. These usually consider likelihood functions from the exponential family such as the Laplace, Poisson, beta and gamma distributions \cite{ggp}. A well-known example is the GP classification model, where the classes are represented by the output of an activation neuron into which a latent GP is fed. 

A slightly different approach to non-Gaussian models, which is not constrained to the exponential family, is the warped GP (WGP, \cite{warped04}). The WGP models non-Gaussian data by assuming that there is  a transformation $\phi$ such that the observations can be \emph{passed through} $\phi$ to yield a GP, therefore, the likelihood function of this model is not designed directly but, rather, induced by the transformation (a.k.a. \emph{warping}). Expressive WGP models can be designed by choosing complex warping functions and covariance kernels; however, this can lead to critical implementation issues for training and prediction.  First, the likelihood function is in general not convex and populated with local maxima, each of which representing a possible model that explains the observations, as a consequence, the use of the standard BFGS method is not guaranteed to find the global maximum unless the appropriate initial condition is provided. Second, predictions under WGP require us to evaluate the inverse warping function, therefore, if this inverse is not available in closed form (as it is the case for the sum of $\tanh(\cdot)$ functions in \cite{warped04}), there is additional computational complexity arising from this numerical approximation. In practice, this can result in an increase of one or two orders of magnitude in prediction times. 

We now address the two issues of WGP identified above, that is, finding appropriate hyperparameters and guaranteeing efficient predictions, by (i) using derivative-free optimisation methods based on the Powell's method and ensemble MCMC, and also (ii) proposing a warping that has known inverse based on the Box-Cox transformation of the Statistics literature.

%% file: background.tex

\section{two drawbacks of  warped GP\lowercase{s}}
    
 \begin{figure*}[ht]
	\includegraphics[width=0.93\textwidth]{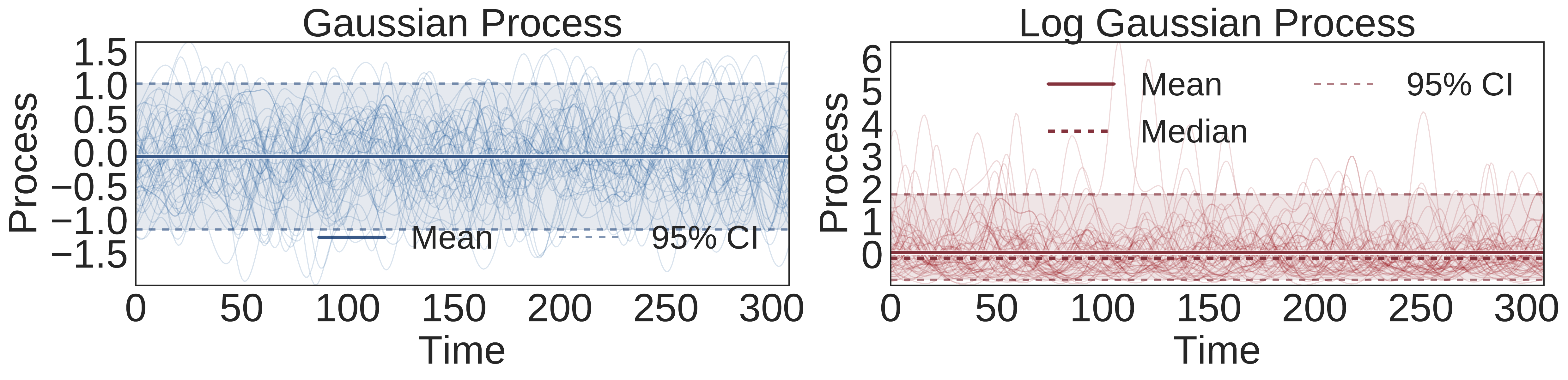}
	\caption{Samples drawn from a GP prior (left) and a logarithmic WGP prior (right). Notice how the log-WGP generates paths that have positive codomain and exhibit large positive deviations (heavy tails).}
	\label{fig:example}
\end{figure*}
 
A Gaussian process \cite{rasmussen06}, denoted by
\begin{equation}
	x\left( t\right) \sim \mathcal{GP}\left( m(t),k\left( t,\bar{t}\right)\right),
 \end{equation}
 is a stochastic process  $\{x_{t}\}_{t\in\mathcal{T}}$ with mean function $m(\cdot)$ and covariance function $k(\cdot,\cdot)$, such that any finite collection of values of the process in $\textbf{t}\in\mathbb{R}^N$ is distributed as a multivariate Normal distribution with mean $m(\textbf{t})$ and covariance $k(\textbf{t},\textbf{t})$. GPs can be used as a building blocks of a non-Gaussian model, one way of doing this is by following the \emph{warping} rationale introduced in \cite{warped04}. 

 A warped Gaussian process is a stochastic process $y$ such that, $\phi(y)=x\sim\GP(m,K)$, where $\phi$ is referred to as the \emph{warping} and $x$ as the \emph{base} GP. A key property of the WGP model for regression is that the predictive and prior distributions belong to the same family: they are both $\phi$-warped Gaussians with known statistics---this closure under conditioning property is inherited from the base GP.

The change of variables Theorem \cite{tao2011introduction} can be used to calculate conditional densities of transformed Gaussian random vectors: For two jointly-Gaussian vectors $\x,\x'$ with conditional density $p(\x\vert \x')=\cN\left( \mu_{\x|\x'} , \Sigma_{\x|\x'}\right)$, and a pair of vectors $\y,\y'$ such that $\x=\phi(\y)$ and $\x'=\phi(\y')$, the conditional density $p(\y\vert\y')$ is given by  
\begin{eqnarray*}
	p\left( \y|\y'\right) &=&\prod\limits_{i=1}^{n}\frac{d\phi\left( y_{i}\right) }{dy} \cN\left( \phi\left( \y\right) | \mu_{\x|\x'} , \Sigma_{\x|\x'}\right) \\
	\mu_{\x|\x'} &=& \mu_{\x}+\Sigma _{\x\x'}\Sigma _{\x'\x'}^{-1}\left( \phi\left( \y'\right) -\mu_{\x'}\right) \\
	\Sigma_{\x|\x'} &=& \Sigma _{\x\x}-\Sigma _{\x\x'}\Sigma _{\x'\x'}^{-1}\Sigma _{\x'\x}
\end{eqnarray*}
where $\Sigma _{\x\x'}$ denotes the covariance between $\x$ and $\x'$, and $\mu_\x$ denotes the marginal mean of $\x$. 

As mentioned above, observe that the posterior density of the transformed element $p\left( \y|\y'\right)$ belongs to the same family as the unconditional density $p(\y)$; this property of closure under conditioning is inherited from the (base) Gaussian pdf and preserved by the coordinate-wise transformation $\phi$. Furthermore, the non-Gaussian multivariate distribution $p(\y)$ is also closed under marginalisation and permutation, again since $\phi$ is defined coordinate-wise. 

 By virtue of the change of variables Theorem \cite{tao2011introduction}, the training and prediction expressions of WGP are stratighforward to derive, however, they are problematic to implement. We discuss these issues in the following two subsections.

\subsection{Model fitting via optimisation and local minima} 
\label{sub:fitting}
In order to construct a general WGP model, it seems that complex forms for both the covariance function $k$ and the transformation $\phi$ are a necessary condition, however, these expressive forms are defined by a large number of parameters. Although this rationale undoubtedly leads to more flexible generative models, the overparametrisation results, in general, in a negative log-likelihood function with several local minima given by
\vspace{-0.5em}
\begin{align}
\label{eq:NLL}
	\text{NLL} &= -\log p(\y|\theta_x,\theta_\phi)\\
	 &= \frac{n\log(2\pi)}{2}  + \frac{1}{2}\left(\phi(\y)-\mu_{\x} \right)^{\top}\Sigma_{\x \x}^{-1}\left(\phi(\y)-\mu_{\x} \right) \nonumber  \\ 
	  &+ \frac{1}{2} \log \left|\Sigma_{\x \x}\right|-\sum_{i=1}^{n}\log\left(\frac{d\phi(y_{i})}{dy}\right), \nonumber
\end{align}
where $\Sigma_{\x \x}$ and $\mu_{\x}$ are the covariance (kernel) and mean of the base GP respectively. As it is well known for deep structures having several parameters \cite{Goodfellow-et-al-2016}, the gradient-based method fails to escape these minima and therefore more elaborate optimisation methods need to be considered. This fact motivates us to depart from the the standard training method for GPs, the Broyden–Fletcher–Goldfarb–Shanno algorithm (BFGS) \cite{wright1999numerical}, towards methods that do not rely on differentiating the NLL. As we will see in the experimental section, we shown how BFGS is trapped in local minima, whereas derivative-free methods such as Powell \cite{powell1964efficient} and ensemble MCMC \cite{goodman2010ensemble} are able to locate better solutions. 

Powell's method is a derivative-free direction set method that only evaluates the $n$-dimensional cost function, and not its derivatives, to construct a set on $n$ conjugate (orthogonal) directions. The supporting concept in Powell's approach is that the minimum of a positive-definite quadratic form can be found by performing at most $n$ successive line searches along mutually conjugate directions \cite{pardalos2002combinatorial}. Also, this procedure can be applied to non-quadratic functions by adding a new composite direction at the end of each cycle of $n$ line searches. 

The second approach to be used in this work is \emph{ensemble} MCMC. Recall that standard MCMC proceeds by constructing a Markov chain such that samples generated by the chain converge to those sampled from a target density. Within optimisation, MCMC can be used to explore the cost function as if it were a distribution from where one is sampling, to then report the solution simply as the sample with lowest cost (or highest probability). Ensemble MCMC refers to sampling from an $n$-dimensional parameter space by constructing a $kn$-dimensional Markov chain, operating on the product-space of $k$ $n$-dimensional spaces, where normally $k \ge 2n$ \cite{foreman2013emcee}. Due to the distributed-exploration nature of ensemble MCMC, the constructed Markov chain converges faster and (simultaneously) explores several high-probability regions of the likelihood function, as we will see later on experimentally.

\subsection{Closed-form prediction} 
\label{sub:closed_form_prediction}

Inference follows from a corollary of the change of variables theorem that states that the probability (measure) of a set $E$ under the density of $\y$, is equal to the probability of the image of $E$, $\phi(E)$, under the density of $\x$. Conditioning on observed data $\y$, we can express the corollary as
\begin{align*}
	\int\limits_{E}p_{y}\left( y|\y\right) dy =\int\limits_{\phi\left( E \right) }p_{x}\left(x|\y\right) dx=\int\limits_{\phi\left( E \right) }p_{x}\left(x|\x\right) dx
\end{align*}%
where the first identity is due to the change of variables Theorem and the second one due to the deterministic relationship $\x=\phi(\y)$. For different choices of the set $E$ and using the inverse transformation $\phi^{-1}$ we can express the median as
\begin{align}
	\text{median}({y(t)}) &=\phi ^{-1}\left(\text{median}({x(t)})\right)=\phi ^{-1}\left(m(t)\right)\label{eq:pred1}
\end{align}
 and its $p$-percentile confidence intervals as
\begin{align}
	I_{y(t)}^p &=\left[ \phi ^{-1}\left(m(t) -z_{p}\sigma(t)\right), \phi ^{-1}\left(m(t) +z_{p}\sigma(t)\right)\right]\label{eq:pred2}
\end{align}
where $\sigma(t) = \sqrt{k(t,t)}$, $z_{p}$ is the quantile of standard Gaussian (ex. $z_{95} \approx 1.96$) and we used the fact that for a Gaussian $\text{median}(x) = \text{mean}(x)$. 

Sampling the non-Gaussian process is also direct: it is only required to simulate a realisation of the GP and then apply the inverse of the transformation, that is,
\begin{align*}
 x(\t)&\sim\GP(m(\t),k(\t,\t))\\
 y(\t)& =\phi^{-1} \left( x(\t) \right). 	
\end{align*}


%% file: trans.tex

\section{An Explicit-Inverse Warping for WGP\lowercase{s}}

We next propose a warping $\phi$ that allows us to calculate eqs.~\eqref{eq:NLL}-\eqref{eq:pred2} analytically. We achieve this based on the Box-Cox transformation and thus refer to the proposed WGP using the Box-Cox transformation  as the \emph{Box-Cox Gaussian Process} (BCGP).

\subsection{Logarithmic transformation}

A standard approach to transform non-Gaussian positive observations into (approximately) Gaussian ones is to apply the logarithmic function, in the WGP setting this is expressed as $\phi (x)= \exp(x)$, whose derivative and inverse are known explicitly. It is particularly interesting that, within the logarithmic transformation, the mean and covariance of  $y$ are no longer the transformations of the statistics $m$ and $k$, but (more generally) we have that the $n^\text{th}$ moment of the $y$ is given by
\begin{align*}
	\mathbb{E}_{y}\left[ y^{n}\right]  &=\exp \left( nm+\frac{n^{2}k}{2}\right).
\end{align*}
This transformation induces a stochastic process of different nature than $x$, since we emphasise that $y$ is always positive. Fig.~\ref{fig:example} shows samples drawn from a standard GP (left) and a logarithmic WGP (right).

\subsection{\mbox{Beyond log-Gaussian: The Box-Cox transformation}}

\begin{figure}[ht]
	\includegraphics[width=0.48\textwidth]{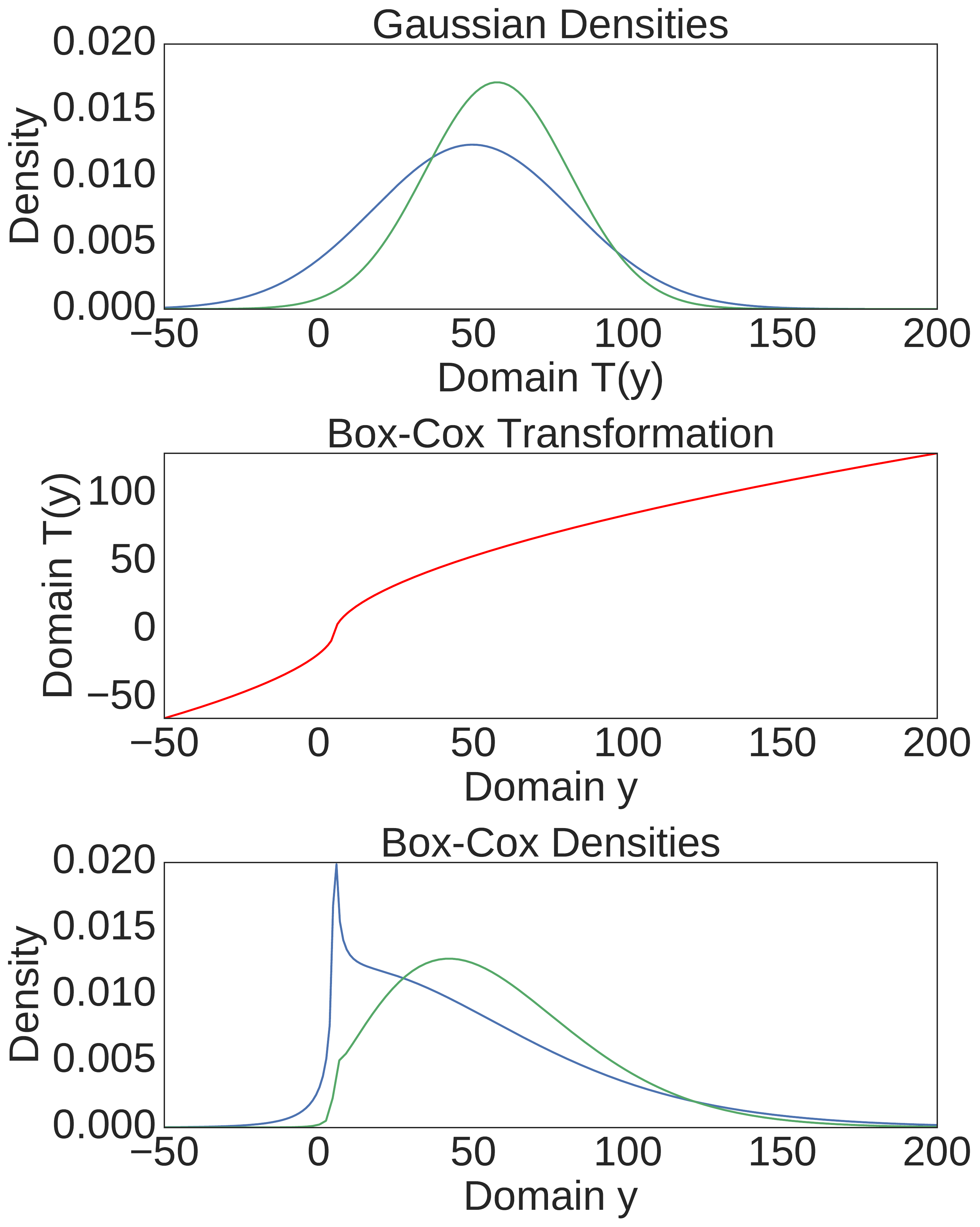}
	\caption{An example of how two Gaussian densities (top), using a Box-Cox transformation (middle), are transformed into two Box-Cox densities (bottom).}
	\label{fig:box_cox_density}
\end{figure}

Although the logarithmic transformation is a standard in the Statistics literature, \cite{boxcox} studies a family of power functions that generalise the logarithm known as \emph{Box-Cox transformations}, these depend on a single parameter $\lambda \in \mathbb{R}^{+}_{0}$ and are given by
\begin{eqnarray}
\label{eq:box-cox-trans}
	\phi _{\lambda }\left( y\right) &=&\frac{sgn\left( y\right) \left\vert
		y\right\vert ^{\lambda }-1}{\lambda } \\
	\frac{d\phi _{\lambda }\left( y\right) }{dy} &=&\left\vert y\right\vert
	^{\lambda -1}\\
	\phi^{-1} _{\lambda }\left( x\right) &=&sgn\left( \lambda x+1\right) \left\vert
	\lambda x+1\right\vert ^{\frac{1}{\lambda }} 
\end{eqnarray}
For $\lambda >0$ the codomain of the transformation is $\mathbb{R}$, whereas for $\lambda =0$ the codomain is $\mathbb{R}^{+}$, since $\lim\limits_{\lambda \rightarrow 0 }\phi _{\lambda}(y) =\log(y)$. The inverse and derivative of this  transformation are known explicitly, therefore, training and prediction using the WGP model induced by the Box-Cox transformation can be performed in an analytic manner. Recall that to provide sound point predictions, non-Gaussian models need to report more than the mean, in this sense, the Box-Cox transformation results in closed-form expressions \cite{powernormal} for the confidence intervals, median, and mode given by 
\vspace{-0.5em}
\begin{align*}
\text{mode}(y)=\left[ \frac{1}{2}\left( 1+\lambda m+\sqrt{\left( 1+\lambda m\right)^{2}+4k\lambda \left( \lambda -1\right) }\right) \right] ^{\frac{1}{\lambda }}.
\vspace{-0.5em}
\end{align*}

Figure ~\ref{fig:box_cox_density} shows an example of the Box-Cox transformation with $\lambda=0.58$ (in red), two base Gaussian densities (in blue and green) and the transformed non-Gaussian densities (in blue and green, respectively).

\subsection{Relationship to other methods}

The original warped Gaussian processes \cite{warped04} considers the monotonic warping of the identity function  given by 
\vspace{-0.5em}
\begin{align}
	\label{eq:wgp}
	\phi \left( y\right)   = y+\sum\nolimits_{j=1}^{J}a_{j}\tanh \left( b_{j}\left(
	y+c_{j}\right) \right)
\end{align}
\vspace{-0.2em}where $a_{j},b_{j}  \geq  0\ \forall j=1,\ldots,J$. This idea has been further extended by considering a transformation given by a single GP with the identity as mean function (Bayesian WGP, \cite{bayesianwarped12}) or a concatenation of GPs (deep GP, \cite{deep_GP}). These extensions point in the direction of providing a flexible warping, however, this comes with a considerable cost for performing predictions, where it is needed to compute the inverse warping. For instance, although the standard WGP \cite{warped04} is the most computationally-efficient of the WGP family, approximating the inverse of $\phi$ in eq.~\eqref{eq:wgp} using the Newton-Raphson \cite{atkinson2008introduction} method increases the prediction running time in one or two orders of magnitude in practice. Additionally, for the Bayesian and deep variants of WGP, besides the prediction cost, there is also a training cost associated to the use of variational inducing variables to deal with the intractability of the model \cite{titsias2009variational}.

All these expressive models rule out standard transformations (WGP) or have a considerable computational complexity in the general case (Bayesian and deep WGP). Conversely, the proposed model is based on a standard transformation from the Statistics literature and provides efficient prediction due to the existence of an analytical inverse warping. 

\subsection{\mbox{More-expressive Box-Cox transformations via compositions}}

The \emph{affine transformation} is given by 
\begin{equation}
\label{eq:affine-trans}
\phi_\text{affine} (y) = a+by,\ \ a,b\in\R	
\end{equation}
and is referred to as \emph{shift} when $b=1$ and as \emph{scale} when $a=0$. The affine transformation does not provide enhanced modelling ability over standard GPs since an affine-transformed GP is still a GP with a shifted mean and scaled variance. However, the affine warping will be composed with Box-Cox transformation to produce more expressive transformations, motivated by the fact that the inverse and derivatives of function compositions are given by the inverses and derivatives of their component functions. For instance, for a composition $\phi(y) =\phi_{2}(\phi_{1}(y))=x$ the inverse and the derivative are given respectively by 
\begin{eqnarray*}
	\phi ^{-1}\left( x\right) &=&\phi _{1}^{-1}(\phi _{2}^{-1}(x)) \\
	\frac{d\phi \left( y\right) }{dy}&=&\frac{d\phi_{2}\left( \phi_{1}\left( y\right) \right) }{dy}\frac{d\phi_{1}\left( y\right) }{dy}.
\end{eqnarray*}

As mentioned above, when the data are strictly positive a standard practice is to apply the logarithmic transformation. Critically, if the data is known to be lower-bounded by an unknown quantity, one can compose the logarithmic transformation with the shift transformation in eq.~\eqref{eq:affine-trans} in order to find the shift parameter during training. An upper bound to the data can be found in an analogous way by replacing the shift by an affine transformation, thus allowing a negative scaling. In this sense, composing two affine-logarithmic allows us to find the upper and lower bounds simultaneously. 

To further relax the strict (lower) bound condition of the logarithmic transformation to a more permissive one, we can also replace the logarithm by the Box-Cox transformation in eq.~\eqref{eq:box-cox-trans}, where the permissiveness of the bound is controlled by the $\lambda$.

\begin{figure}[ht]
	\includegraphics[width=0.48\textwidth]{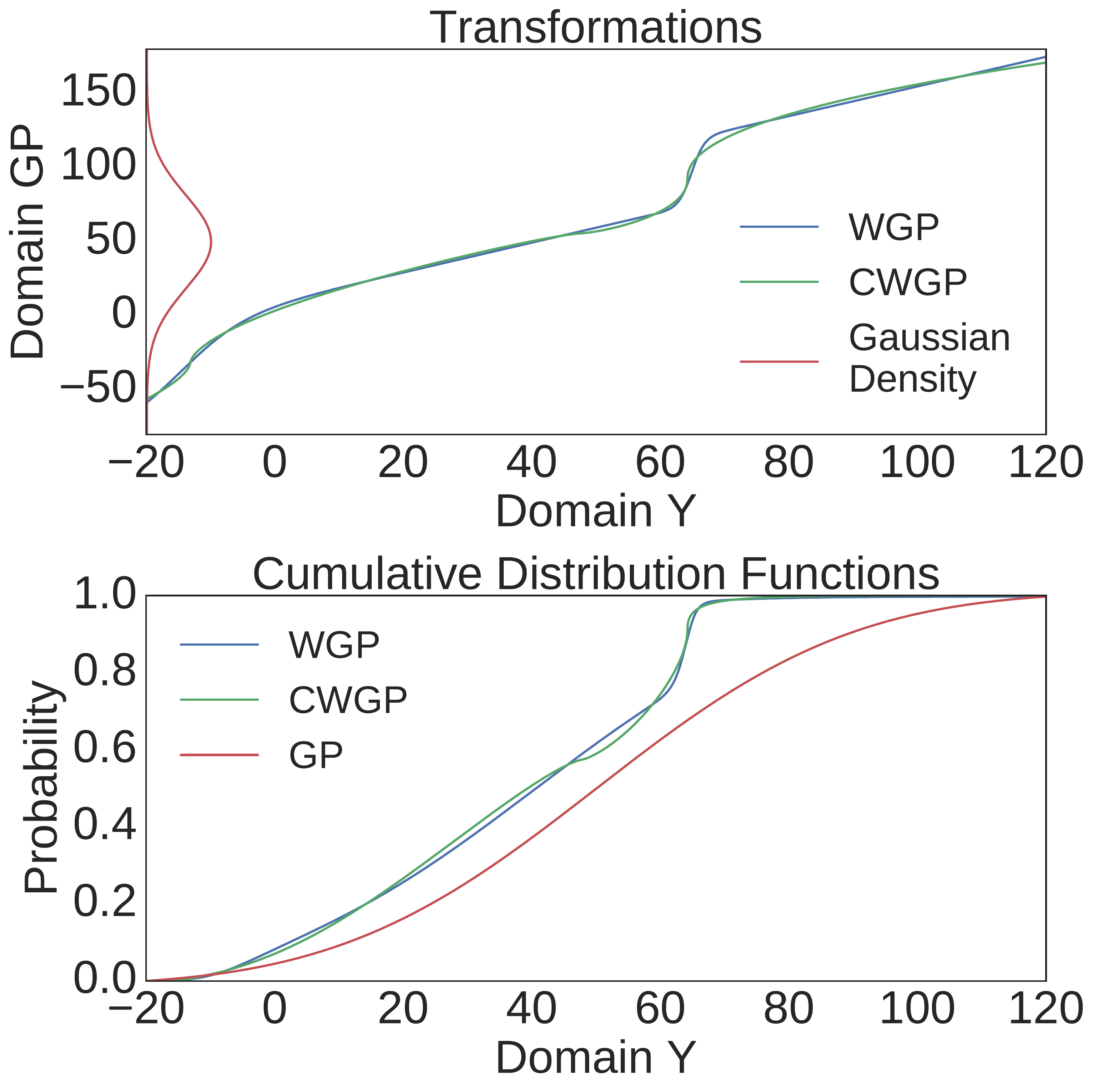}
	\caption{Approximating a WGP warping (sum of hyperbolic tangents, blue) using the proposed compositional warping (Box-Cox and Affine functions, green). The top plot shows the Gaussian density, the WGP transformation and the (proposed) BGCP approximation, whereas the bottom plot shown all CDFs: Gaussian, WGP and BCGP approximation of WGP.}
	\label{fig:cwgp_flexible}
\end{figure}

We now show that the proposed compositional warping can replicate an usual warping architecture implemented by WGP, that is, a sum of two hyperbolic tangents as in eq. \eqref{eq:wgp}. We approximated this warping with a composition of two Box-Cox and affine transformations fitted via least squares. Fig. \ref{fig:cwgp_flexible} shows the $\tanh(\cdot)$ (blue) and our Box-Cox composition (green) warpings and induced distributions. Observe the point-wise similarity of the warpings and probability distributions.

\subsection{Approximating the moments} 

To perform predictions, it is often necessary to compute the (posterior) moments of the marginal distribution of WGP, this involves numerical approximations. Relying on the change of variables theorem, we can formulate the expectation of a measurable function $h:\mathcal{Y}\rightarrow \mathbb{R}$ under the non-Gaussian law $p(\y)$ as an expectation under the Gaussian law $p(\x)$ given by
\begin{align*}
	\mathbb{E}_{\y}\left[ h\left(\y\right)\right]  =\mathbb{E}_{\x}\left[h\left(\phi ^{-1}\left( \x\right)\right) \right]
\end{align*}
which can be efficiently computed numerically using the Gauss-Hermite quadrature \cite{gausshermite64}, for which $k$-point approximations are exact when the integrand $h \circ \phi^{-1}$ is a polynomial of order $2k-1$ or less. Choosing $h(y)=y$, the approximation of the mean of $y$ is given by
\begin{eqnarray}
	\mathbb{E}_{y}\left[ y\right]  &=& \int\limits_{  }\phi ^{-1}\left( x \right) p_{x}\left(
	x\right) dx \nonumber\\
	 &\approx & \frac{1}{\sqrt{\pi}} \sum\limits_{i=1}^{k} w_{i}\phi ^{-1}\left( \sqrt{2}\sigma_{x} x_{i} + m_{x} \right) \label{eq:exp_CWGP}
\end{eqnarray}
where the weights $\{w_{i}\}_{i=1}^k$ and locations $\{x_{i}\}_{i=1}^k$ are given by the Gauss-Hermite quadrature method \cite{gausshermite64}. Obtaining a formula to approximate the variance is analogous.

Finally, observe that evaluating $\phi^{-1}$ is required to compute expectations, the median and confidence intervals of the non-Gaussian model. Since for our model $\phi^{-1}$ is known, the cost of evaluating it is $\O(d)$, where $d$ is the number of Box-Cox and affine components of $\phi$. Therefore, the cost of evaluating $\mathbb{E}_{y}\left[ y\right]$ in eq. \eqref{eq:exp_CWGP} using the $k$-point Gauss-Hermite quadrature is $\O(kd)$ for our model, which give computationally-efficient approximations due to the polynomial nature of the Box-Cox transformation. Conversely, WGP approximates $\phi^{-1}$ using the Newton-Raphson  method \cite{atkinson2008introduction}, meaning that the cost of evaluating $\mathbb{E}_{y}\left[ y\right]$ for WGP is $\O(kdt)$, where $t$ is the number of iterations of Newton-Raphson. In practice, the availability of an explicit expression for $\phi^{-1}$ makes our proposed model between one and two orders of magnitude more computationally efficient than WGP. 

\begin{figure*}[ht]
	\includegraphics[width=0.48\textwidth]{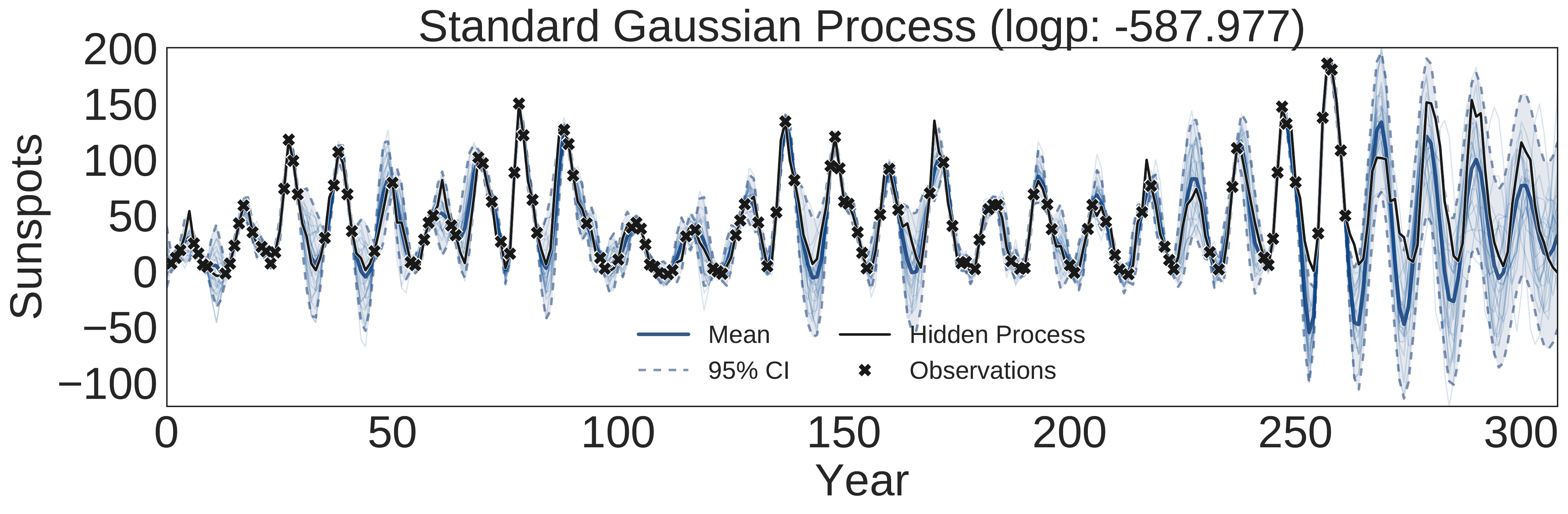}
	\includegraphics[width=0.48\textwidth]{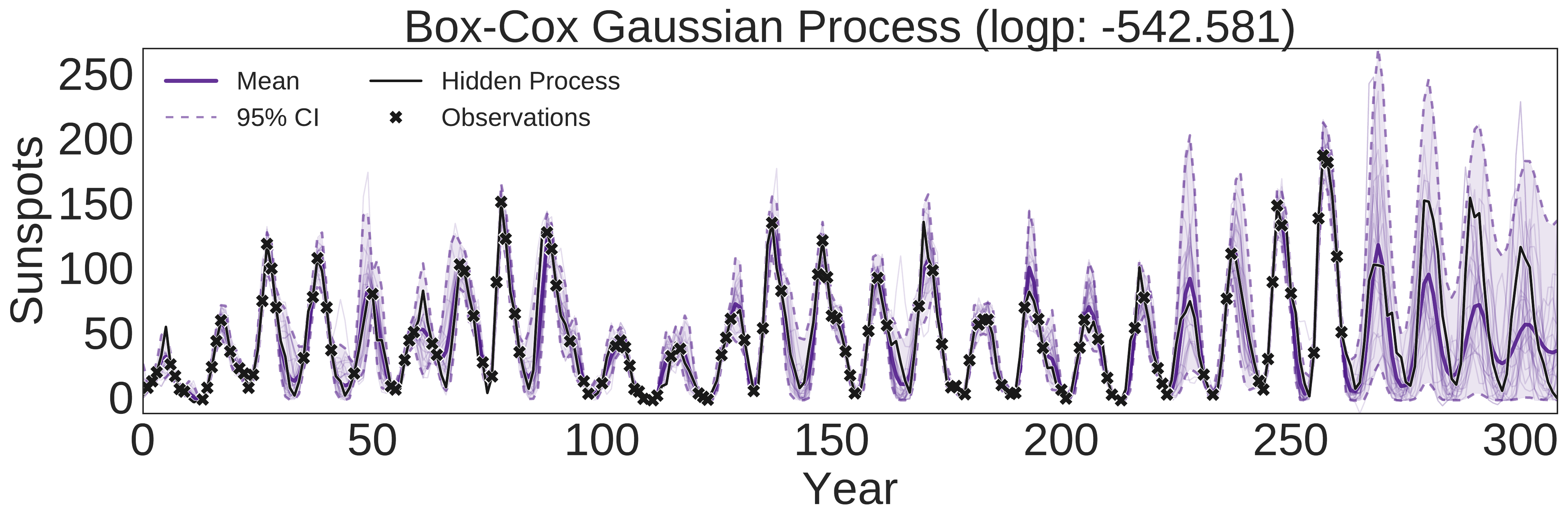}
	\caption{Reconstruction and forecasting of the Sunspot series using GP (top) and BCGP (bottom) trained using BFGS-Powell. The log-likelihood is shown with the titles. Notice the (incorrect) symmetry of the GP posterior and the skewed posterior found by BCGP.}
	\label{fig:sunspotGP_rec}
\end{figure*}

%% file: exper.tex

\section{Simulations: An accurate training framework for the Box-Cox GP}

We validated the proposed Box-Cox GP in two parts: the first one illustrates the advantages of derivative-free training for the  Box-Cox GP and uses it for reconstruction and forecasting of a Sunspot time series. The second part presents a routine of model average and selection of the proposed model using ensemble MCMC on macroeconomic data, where the modes of the solutions are found using a Dirichlet process \cite{blei2006variational}. We emphasise that, as our aim is to construct computationally efficient models, comparisons against Bayesian and deep GPs is out of the scope of this paper. 
\subsection{Performance indices} 
\label{sub:indices}

The models implemented were evaluated via three performance indices which should be interpreted as \emph{the lower the better}. For a test set $\{y_i\}_{i=1}^n$ and reported predictive means $\{y^*_i\}_{i=1}^n$, we considered the point-prediction measures given by the mean squared error (MSE) and the mean absolute error (MAE)---respectively:
\begin{align*}
	\text{MSE} = \frac{1}{n}\sum_{i=1}^{n}(y_{i} - y_{i}^{*})^2 \text{\quad  and\quad  }
	\text{MAE} = \frac{1}{n}\sum_{i=1}^{n}|y_{i} - y_{i}^{*}| 
\end{align*}
and also the negative log prediction distribution (NLPD), a measure of distribution prediction error given by
\begin{align}
	\label{eq:NLPD}
	\text{NLPD} &=-\frac{1}{n}\sum_{i=1}^{n}\log(p_{i}(y_{i})).
\end{align}

\subsection{Reconstruction and forecasting of the Sunspots time series} 
\label{sub:bfgs_train}


We first considered the Sunspot time series available from the \texttt{scikit-learn} toolbox in Python. This time series represents the yearly number of sunspots for each year between 1700 and 2008, with a total of 309 data points. We considered 131 observations for training, these were randomly selected only between years 1700 and 1961 (50\% of missing data), this allowed us to perform two experiments: reconstructing the signal between years 1700 and 1961, and forecasting the signal from year 1961 to 2008 (47 data points). The ability of the proposed BCGP for reconstruction and forecasting was compared to the standard GP via the MAE, MSE and NLPD scores on the corresponding test sets.
 
 Both the GP and BCGP models used a 2-component spectral mixture (SM) kernel \cite{Wilson:2013,MOSM}, which is particularly difficult to adjust since it is equivalent to fitting a Gaussian RBF to the (sample approxiation of the) power spectral density of the process; this leads to a NLL with several local minima. Each model was trained minimising the NLL in eq.~\eqref{eq:NLL} using two strategies starting from a common initial point: the derivative-based BFGS method \cite{wright1999numerical}, and an iterative procedure that implemented  BFGS and the derivative-free global optimisation Powell \cite{powell1964efficient} sequentially; we refer to the second method as BFGS-Powell. This resulted in two sets of hyperparameters for each model (GP and BCGP), thus yielding four trained models.
 
Table \ref{tab:table_sunspots} shows the scores of all four trained models both for the reconstruction and the prediction examples. For all indices, we can see that the proposed training procedure (BFGS-Powell) succeeded in unlocking the potential of BCGP to discover non-Gaussian features, where the  BFGS-trained BCGP is only marginally better than a standard GP. Critically, even for the standard GP, BFGS-Powell improved over the pure BFGS model. Fig.~\ref{fig:sunspotGP_rec} shows the reconstruction and forecasting of the series using both the best GP (top) and the best BCGP (bottom). Notice how both models successfully learnt the right frequency, which is the difficult part when using the SM kernel, however, the GP fails to estimate the range of the time series, while BCGP learns that the data are always positive and have a skewed marginal density
 
 \begin{table}
	 \vspace{-0.5em}
 	{\small
 		\caption{Performance of GP and BCGP for reconstruction and forecasting of the Sunspots data trained using BFGS and BFGS-Powell.}
 		\label{tab:table_sunspots} 
 	}
 	\centering
 	\begin{tabularx}{250pt}{clrrrr}
 		&{} &         MAE &           MSE &      NLPD &         NLL \\
 		\hline
 		\hline
 		&GP BFGS &       11.06 &      237.19 &        4.06 &      608.27 \\
 		\raisebox{-.5\normalbaselineskip}[0pt][0pt]{\rotatebox[origin=c]{90}{Reconst.}} &GP BFGS-Powell  &       10.37 &      217.96 &        4.03 &      587.98 \\
 		&BCGP BFGS &       11.06 &      239.36 &        4.03 &      578.68 \\
 		&BCGP BFGS-Powell  &        \textbf{8.85} &      \textbf{150.36} &        \textbf{3.90} &      \textbf{542.58} \\
 		\hline
 		&GP BFGS  &       40.36 &     2509.55 &        5.36 &      608.27 \\
 		\raisebox{-.5\normalbaselineskip}[0pt][0pt]{\rotatebox[origin=c]{90}{Forecast}} &GP BFGS-Powell   &       30.68 &     1414.81 &        5.17 &      587.98 \\
 		&BCGP BFGS  &       40.25 &     2526.24 &        5.20 &      578.68 \\
 		&BCGP BFGS-Powell   &       \textbf{26.90} &     \textbf{1253.10} &        \textbf{4.95} &      \textbf{542.58} \\
 		\hline
 		\hline
 	\end{tabularx}
 	\vspace{-2.0em}
 \end{table}

\subsection{Learning a macroeconomic time series} 
\label{sub:mcmc_train}

\begin{figure}[ht]
	\includegraphics[width=0.48\textwidth]{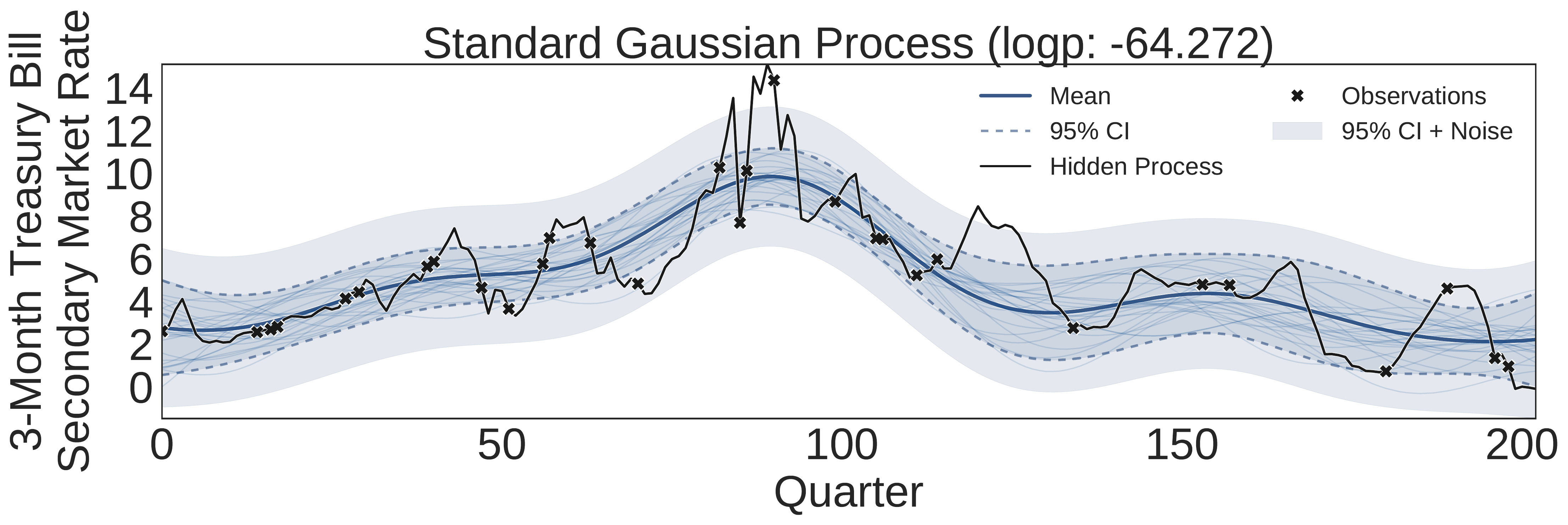}
	\includegraphics[width=0.48\textwidth]{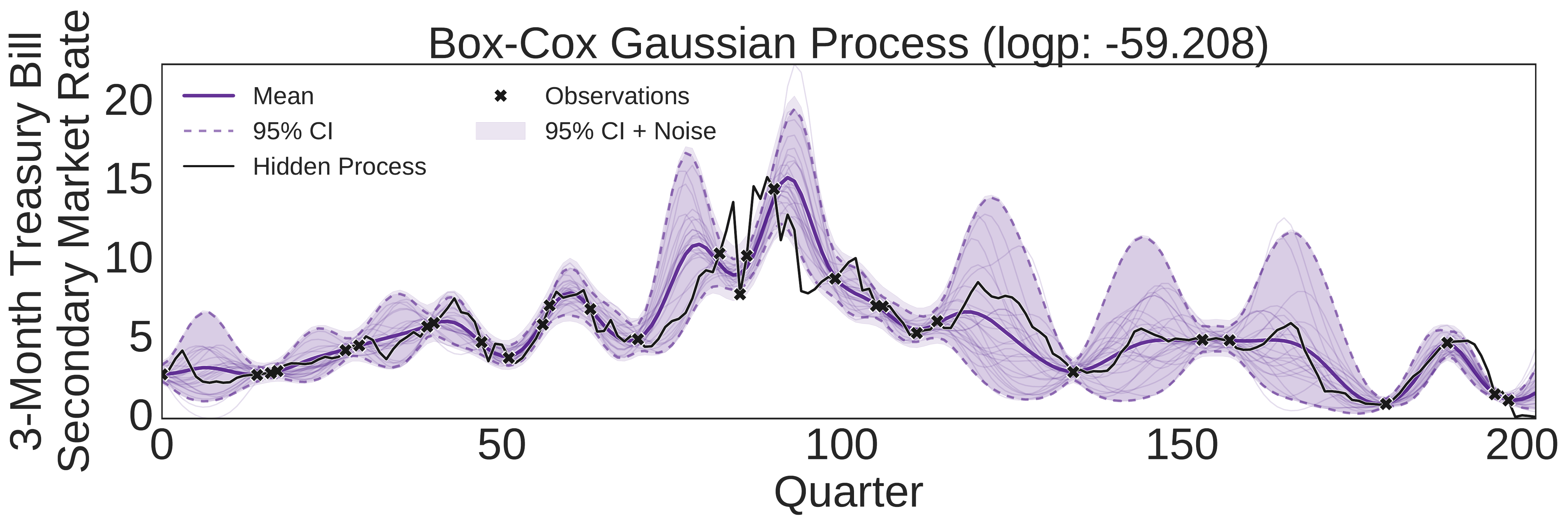}
	\caption{Standard GP (top) and Box-Cox GP (bottom) trained using the BFGS-Powell method on a macroeconomic time series.}
	\label{fig:bfgs_powell}
\end{figure}

\begin{figure}[ht]
	\includegraphics[width=0.48\textwidth]{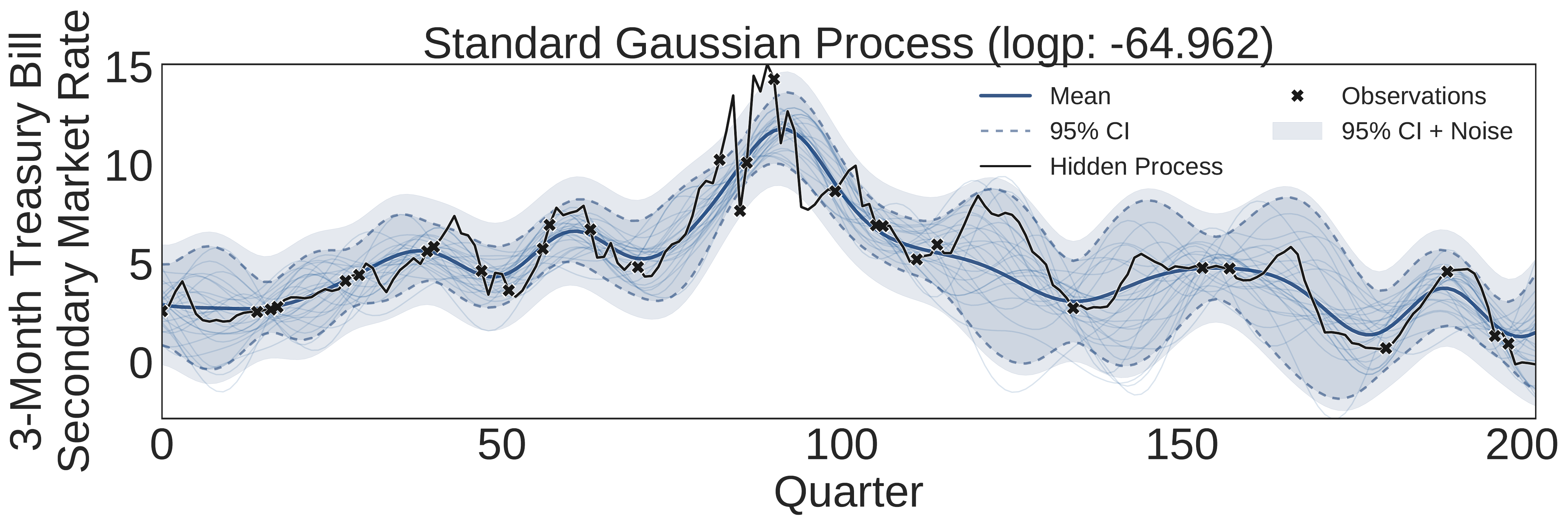}
	\includegraphics[width=0.48\textwidth]{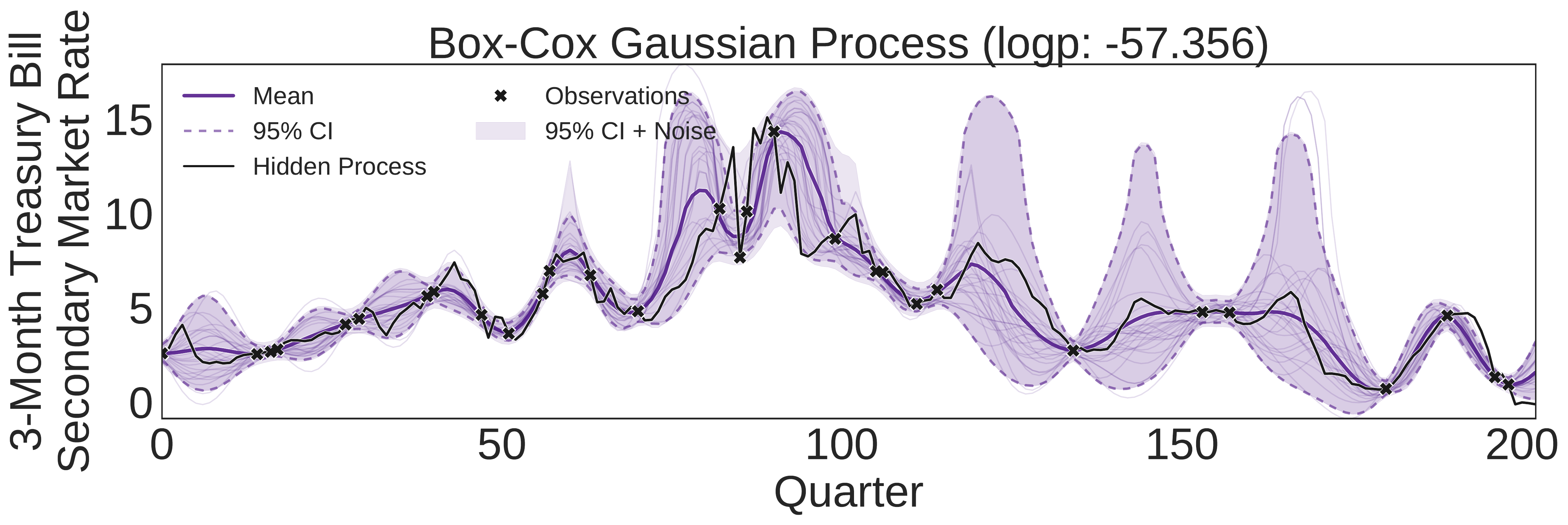}
	\caption{Standard GP (top) and Box-Cox GP (bottom) trained using the ensemble MCMC method on a macroeconomic time series.}
	\label{fig:ensemble_mcmc}
\end{figure}

\begin{figure*}[ht]
	\includegraphics[width=0.98\textwidth]{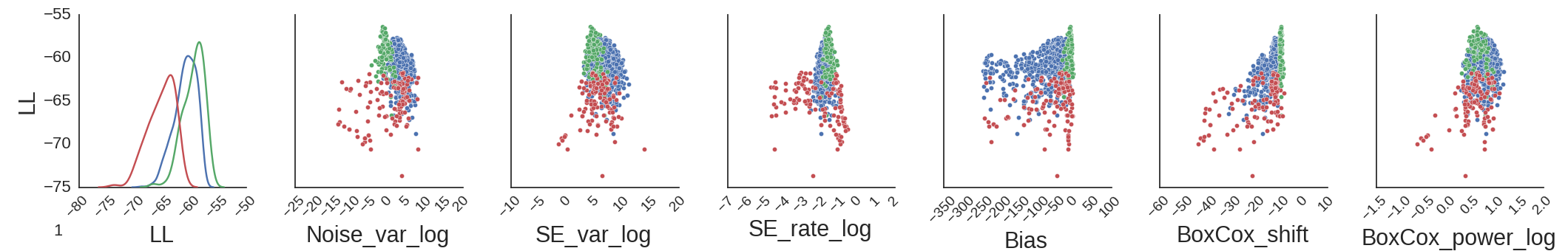}
	\caption{Scatter plot of BCGP hyperparameters against their log-likelihood using with ensemble MCMC for the macroeconomic data. The first plot is a smoothed histogram and the colours denote the three clusters found using the Bayesian Gaussian mixture model.}
	\label{fig:tgp_params}
	\vspace{-1em}
\end{figure*}

We next considered the quarterly average \emph{3-Month Treasury Bill: Secondary Market Rate} \cite{tb3ms}, representing the price of U.S. government risk-free bonds, which cannot take negative values and can have large positive deviations. Out of the 203 observations between 1959 and 2009, we randomly (uniformly) selected 30 datapoints (the 15\% of the data) for training both GP and BCGP, and left the remaining 85\% for evaluation. Fig.~\ref{fig:bfgs_powell} shows the GP and BCGP models trained using the BFGS-Powell procedure described in the previous section. Notice how the standard GP fails to adjust to the data, misidentifies the (zero) lower bound, and shows excessive noise variance---the proposed BCGP did not suffer of any of these issues. 

\begin{figure}[ht]
	\includegraphics[width=0.48\textwidth]{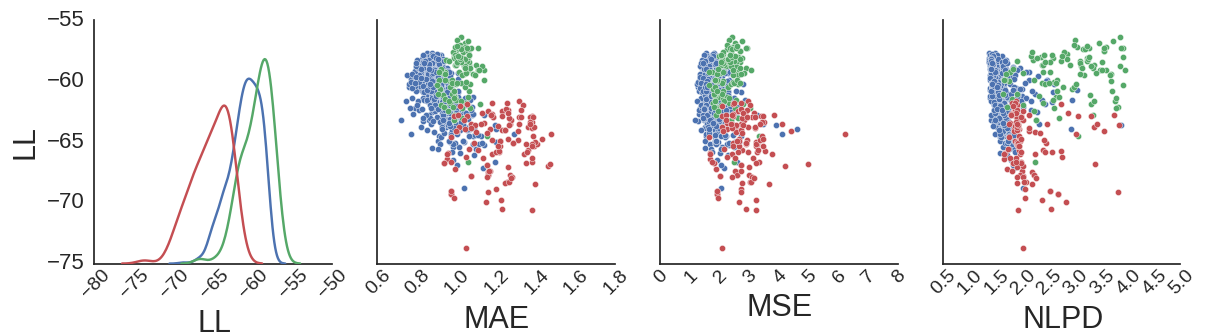}
	\caption{Scatter plot of log-likelihood against scores for BCGP trained on macroeconomic data.}
	\label{fig:tgp_errors}
\end{figure}

\begin{figure}[ht]
	\includegraphics[width=0.48\textwidth]{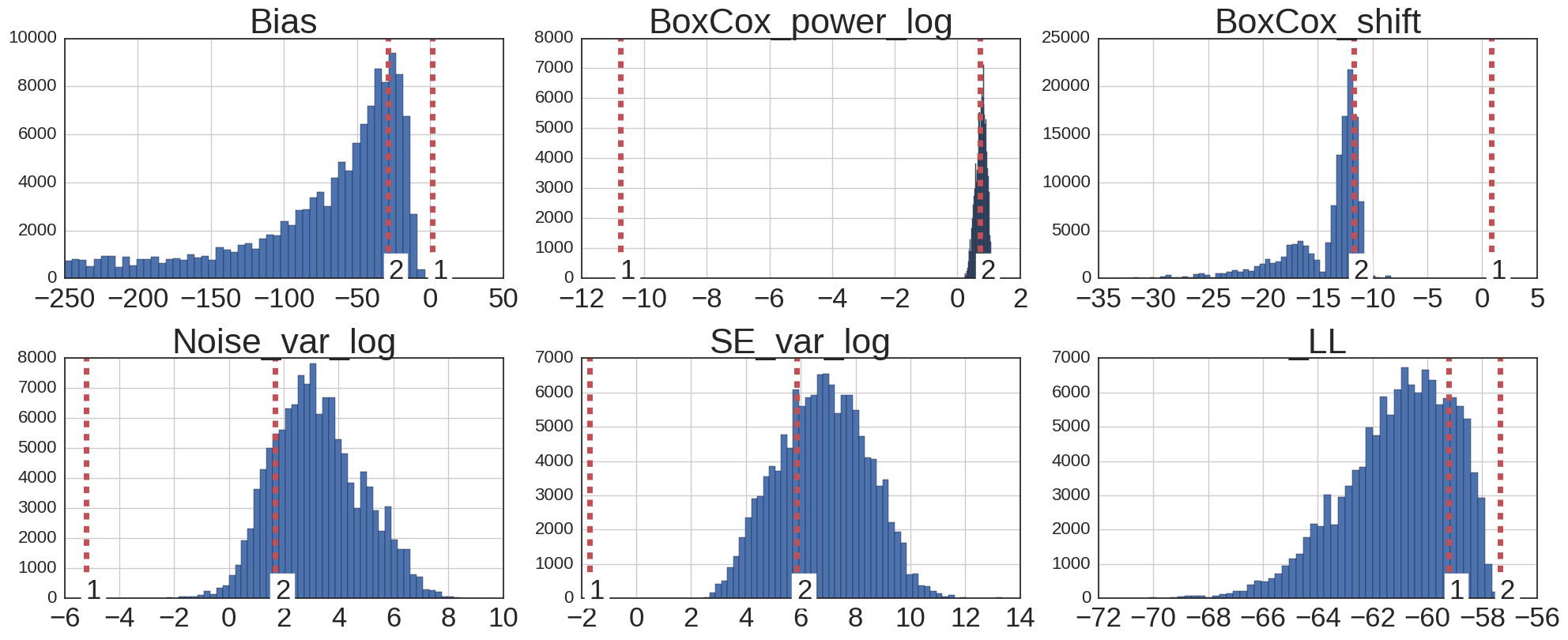}
	\caption{BCGP marginals of likelihood: line 1 denote the BFGS-Powell selected and line 2 is the ensemple MCMC model selected.}
	\label{fig:models_maps}
\end{figure}

We then found the full posterior of the model hyperparameters using ensemble MCMC \cite{goodman2010ensemble} (using an uninformative prior). Fig.~\ref{fig:ensemble_mcmc} shows the GP and BCGP models trained using the ensemble MCMC procedure described in the previous section. Fig.~\ref{fig:tgp_params} shows a scatter plot of all hyperparameter samples against their marginal likelihood, where we used the \textit{Bayesian Gaussian Mixture Model} \cite{blei2006variational} to find the number of clusters of the posterior samples. This revealed the existence of three-well defined modes for the model likelihood, were the classic maximum-likelihood (ML) solution is found in the green mode. Fig.~\ref{fig:tgp_errors} also show all the scores against the log-likelihood colour-coded per mode, where unlike the ML criteria, the mode that has lowest scores (MAE, MSE and NLPD) is shown in blue.
Fig.~\ref{fig:models_maps} shows a histogram of the hyperparameters in the blue cluster, where the model selected by BFGS-Powell is marked with the symbol ``1'' and the ML hyperparameters found via MCMC with the symbol ``2''. Note that the MCMC solution is much closer to the marginal modes than that of BFGS-Powell.

Finally, Table~\ref{tab:table_macroeconomic} shows the scores of the chosen models. Notice the critical improvement of the MCMC model selection against BFGS-Powell, and that the proposed model was able to find an accurate probabilistic representation for the non-Gaussian behaviour of the time series.

\begin{table}[ht]
	{\small
		\caption{Performance of GP and BCGP for reconstruction of macroeconomic data trained using BFGS-Powell and ensemble MCMC.}
		\label{tab:table_macroeconomic} 
	}
	\centering
	\begin{tabularx}{250pt}{lrrrr}
		\hline
		{} &         MAE &           MSE &      NLPD &         NLL \\
		\hline
		GP BFGS-Powell  &        1.28 &        2.83 &        1.94 &       64.27 \\
		GP MCMC &        0.95 &        1.69 &        1.74 &       64.96 \\
		BCGP BFGS-Powell  &        0.93 &        1.94 &        1.69 &       59.21 \\
		BCGP MCMC &        \textbf{0.88} &        \textbf{1.75} &        \textbf{1.42} &       \textbf{57.36} \\
		\hline
	\end{tabularx}
\end{table}

%% file: disc.tex

\section{Discussion}

We have proposed a prediction-efficient warped GP model based on the Box-Cox transformation. This transformation has analytical inverse and a polynomial nature, thus allowing for the design of expressive non-Gaussian models while relying on minimal numerical approximations for prediction. The proposed model has been paired with a novel training procedure using derivative-free methods, namely  Powell and ensemble MCMC, to avoid the optimiser to become trapped in local minima. Through two case studies using real-world data, the proposed Box-Cox GP has exhibited the ability to discover non-Gaussian features from observed data, reconstruct and forecast time series, show the shortcoming of gradient-based methods and illustrate the appealing performance of Powell and the ensemble MCMC for WGP training. Finally, further research in transformed GPs will be devoted to constructing even more expressive transformations, for instance, concatenating multiple warpings having analytical inverse and also multi-coordinate transformations such as the Gaussian process mixture of measurements \cite{gpmm17}. 